\documentclass[11pt]{article}

\usepackage{acl}
\usepackage{times}
\usepackage{latexsym}

\usepackage[T1]{fontenc}

\usepackage[utf8]{inputenc}

\usepackage{microtype}

\usepackage{inconsolata}

\usepackage{graphicx}
\usepackage{natbib}
\usepackage{amsmath}
\usepackage{amssymb}
\usepackage{booktabs}
\usepackage{colortbl}
\usepackage{xcolor}
\usepackage{multirow}
\usepackage{placeins}
\usepackage{url}
\usepackage{float}
\usepackage{capt-of}
\definecolor{headerblue}{RGB}{220, 230, 242}
\definecolor{ourspink}{RGB}{255, 228, 225}
\usepackage{subcaption}
\usepackage{pifont}
\usepackage{caption}
\usepackage{multicol}
\newcommand{\circled}[1]{\textcircled{\scriptsize #1}}
\title{
EMMA: Event-Driven Multimodal Memory System with \\ Multi-Agent Retrieval}

\author{
  \textbf{Ruochen Zhao\textsuperscript{1,3}\thanks{Equal contribution.}},
  \textbf{Jiazheng Kang\textsuperscript{1,3}\footnotemark[1]},
  \textbf{Gezi Li\textsuperscript{2}},
  \textbf{Dongying Lin\textsuperscript{2}}, \\
  \textbf{Ting Bai\textsuperscript{1,3}\thanks{Corresponding author.}}
\\
\\
  \textsuperscript{1}Beijing University of Posts and Telecommunications \\
  \textsuperscript{2}Huawei Technologies Co., Ltd. \\
  \textsuperscript{3}BaiJia AI Team
\\
  \small{
    \texttt{\{kjz,zhaorc,baiting\}@bupt.edu.cn}
  } \\
  \small{
    \texttt{\{ligezi,lindongying\}@huawei.com}
  }
}

\begin{document}
\maketitle
\begin{abstract}
Long video understanding requires reasoning over hours of content that far exceed the context window of multimodal large language models (MLLMs). Existing approaches compress video into fixed-length clips, yet events are fragmented and unrelated content is merged, yielding noisy retrieval. To overcome this challenge, we propose \textbf{EMMA}, an \textbf{E}vent-driven \textbf{M}ultimodal \textbf{M}emory system with Multi-\textbf{A}gent retrieval, which organizes memory around semantic events in three stages: (1) Event segmentation: a dual-stream detector adaptively locates event boundaries by fusing visual semantic drift with spoken transitions, rather than fixed windows. (2) Memory construction: built on the detected events, a three-level hierarchy of raw video, clip-level memory, and event-level memory stores content at progressively higher semantic abstraction. (3) Memory retrieval: a multi-agent framework collaboratively queries this hierarchy through query decomposition, tiered search, evidence validation, and answer generation, narrowing evidence from coarse events to precise clips. Extensive experiments on three challenging benchmarks demonstrate substantial improvements, showing the effectiveness of our event-aligned memory system. Code, prompt templates, and reproduction scripts are available at
\url{https://github.com/BAI-LAB/EMMA}.
\end{abstract}

\section{Introduction}
\label{sec:intro}

Long video understanding requires reasoning over hours of visual content
that forms a coherent storyline. Each frame is embedded in a continuous
narrative structure, yet multimodal large language models
(MLLMs)~\cite{liu2023llava, bai2025qwen3} cannot ingest such lengthy
input in a single pass. Existing memory-based
systems~\cite{fan2024videoagent, zuo2025videolucy, long2025m3agent}
compress video into retrievable units, but they share a fundamental design
choice: unit boundaries are decided by predefined temporal parameters
rather than by the video's event structure. This content-agnostic
segmentation fractures the intrinsic storyline, causing two systematic
retrieval failures. First, \emph{evidence fragmentation}: a coherent event
is split across adjacent units, so no single retrieval can reassemble the
complete evidence. Second, \emph{evidence dilution}: an event shorter than
the preset window is merged with unrelated material, degrading semantic
specificity.

Aligning memory units with the video's intrinsic event structure is a
natural remedy, grounded in Event Segmentation Theory
(EST)~\cite{zacks2007event}: the human brain partitions ongoing
experience into discrete, internally consistent event units to enable
precise and efficient recall. Translating EST into a computational
framework imposes three desiderata. \emph{(D1)~Semantic alignment}:
boundaries must respond to content drift, not elapsed time.
\emph{(D2)~Cross-modal robustness}: a boundary may appear visually or
verbally alone, so a single modality is structurally insufficient.
\emph{(D3)~Multi-granularity retrieval}: the QA layer must adaptively
choose evidence granularity, from a short utterance to a whole event
trajectory.

We instantiate these desiderata in \textbf{EMMA}, an
\textbf{E}vent-driven \textbf{M}ultimodal \textbf{M}emory system with
\textbf{M}ulti-\textbf{A}gent retrieval, through three coordinated
components: \textbf{(1)~Event segmentation} (\emph{D1-D2}): a dual-stream
detector fuses visual semantic drift with spoken chapter transitions to
locate event boundaries adaptively, replacing fixed temporal windows.
\textbf{(2)~Memory construction}(\emph{D3}): atop the detected events, a three-level
hierarchy of raw video, clip-level memory (CLM), and event-level memory
(ELM) stores content at progressively higher semantic abstraction.
\textbf{(3)~Memory retrieval} (\emph{D3}): a multi-agent framework collaboratively
queries this hierarchy through query decomposition, tiered search,
evidence validation, and answer generation, narrowing evidence from coarse
events to precise clips. Extensive experiments on three challenging
long-video benchmarks demonstrate that \textbf{EMMA} achieves substantial
improvements over strong baselines, showing the effectiveness of
event-aligned memory organization. The contributions of our work are:
\begin{itemize}
    \item We diagnose evidence fragmentation and evidence dilution caused by content-agnostic boundaries in existing memory-based video agents, and propose dual-stream event segmentation that adaptively locates event boundaries by fusing visual semantic drift
    with spoken chapter transitions through cross-modal validation.
    \item We construct a three-level memory hierarchy (raw video,
    clip-level memory CLM, event-level memory ELM) atop the detected events, storing content at progressively higher semantic abstraction.
    \item We design a multi-agent collaborative retrieval framework that queries the hierarchy through query decomposition, tiered search, evidence validation, and answer generation, and show on three benchmarks that the combined system outperforms the
    strongest baseline by +4.3 / +3.2 / +2.8 points while also
    reducing retrieval rounds.
\end{itemize}

\section{Related Work}

\subsection{Long-Context Video Understanding}

Applying MLLMs to video typically involves sampling or
compressing frames into a single forward
pass~\cite{wang2024qwen2vl, bai2025qwen3, li2024llavaonevision,
chen2024internvl25}. While effective for short clips, this
degrades as duration grows because the finite context budget
reduces per-frame density. Benchmarks such as
Video-MME~\cite{fu2025videomme} and LVBench~\cite{wang2025lvbench}
quantify this degradation, and EgoLife~\cite{yang2025egolife}
extends it to week-long streams.
Two strategies attempt to spend the budget more wisely.
Visual token compression~\cite{he2024malmm, li2024videochatflash,
song2024moviechat} reduces per-frame cost uniformly,
sacrificing detail where it matters most. Keyframe
selection~\cite{wang2024videotree, wang2025adaretake} discards
low-relevance frames entirely, but a pre-computed relevance
score cannot surface brief events in hour-long videos. Both
trade fidelity or recall for throughput.
Crucially, these methods treat video as a flat frame
sequence, ignoring its inherent event structure. Event
segmentation has been studied for offline summarization
and action recognition, yet rarely as a substrate for
interactive memory systems.

\subsection{RAG-based Video Understanding}

A natural response to the fixed-window limitation is to
offload storage to an external memory that is populated
offline and queried at inference time. RAG-based approaches
index textual captions of fixed-length clips and retrieve
the top-$k$ at query time; graph-structured variants
like LightRAG~\cite{guo2024lightrag} encode cross-segment
relations to support more structured lookups. Because the
memory is built before the question arrives, these systems
can scale to arbitrarily long videos without increasing
per-query latency.

The benefit comes with a representational cost: visual
content is converted to text before storage, so retrieval
quality is bounded by what the captioner chose to verbalize.
Rare or visually subtle events---the very cases that
motivate memory-based designs---are the most likely to be
omitted. EgoRAG~\cite{yang2025egolife} organizes egocentric
streams into hourly and daily summaries, but still relies
on predetermined temporal scales. SceneRAG~\cite{zeng2025scenerag}
builds a scene--entity graph over narrative-consistent scenes,
but derives boundaries primarily from ASR transcripts and temporal
heuristics, which can be unreliable when speech is sparse or noisy.
For RAG systems that retain predetermined temporal units, the atomic
memory unit remains a fixed-length segment whose boundaries are
chosen before examining content, leaving the system vulnerable to
the same boundary effects that plague flat context windows.

\begin{figure*}[t]
    \centering
    \includegraphics[width=\textwidth]{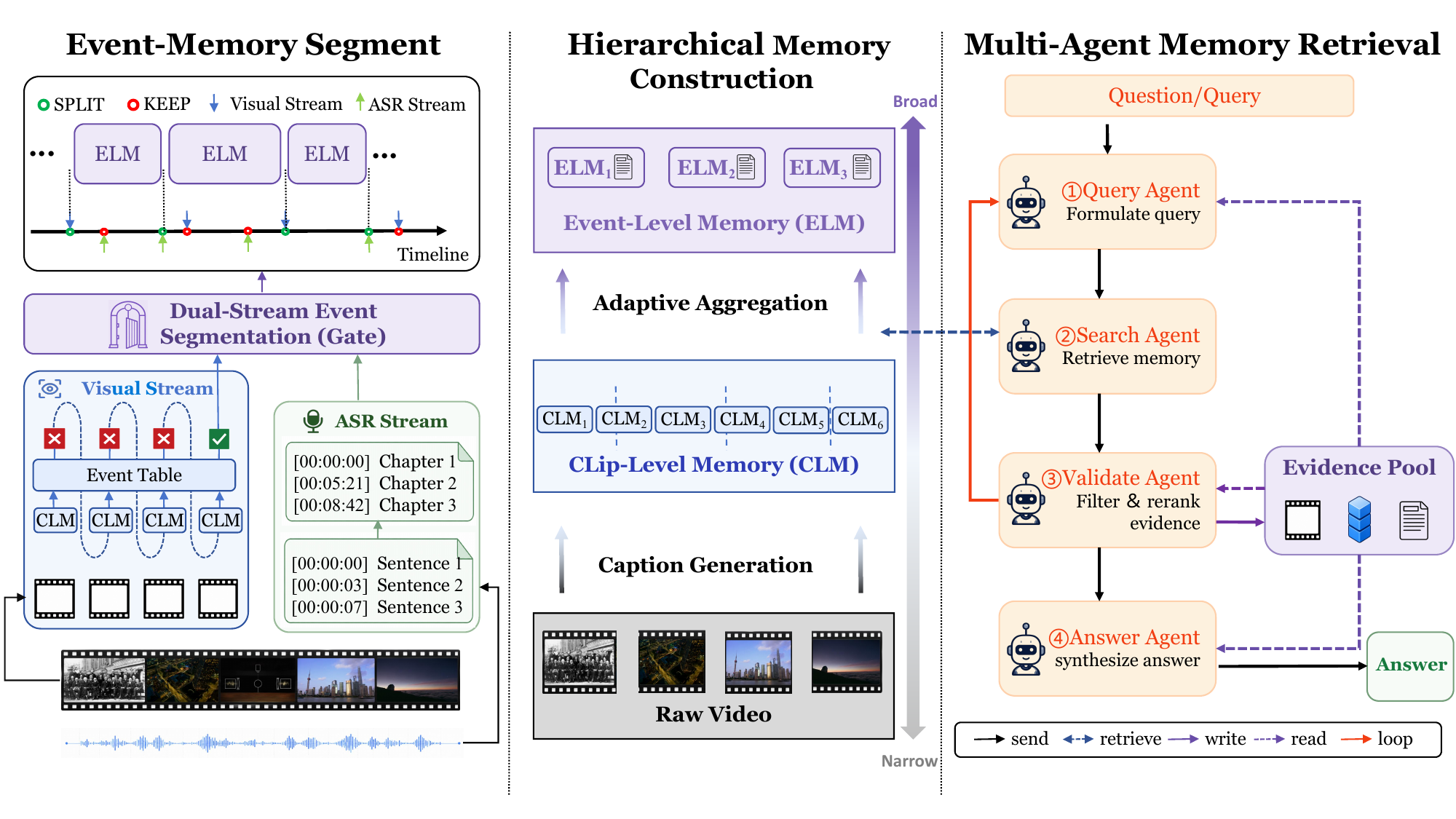}
    \caption{EMMA Framework overview. \textbf{Left:} dual-stream event
    segmentation (visual + ASR) with cross-modal boundary confirmation.
    \textbf{Middle:} three-level memory hierarchy with detected boundaries
    aggregating CLMs into ELMs. \textbf{Right:} four-agent broad-to-narrow
    retrieval loop.}
    \label{fig:overview}
\end{figure*}

\subsection{Memory-based Video Understanding}

More expressive architectures address representational
limits through structured memory. M3-Agent~\cite{long2025m3agent}
maintains an entity-centric multimodal graph for multi-turn
reasoning. WorldMM~\cite{yeo2025worldmm} combines episodic,
semantic, and visual memories with adaptive multi-scale
retrieval. VideoLucy~\cite{zuo2025videolucy} constructs a
three-level caption hierarchy with agent-driven backtracking.
Ego-R1~\cite{tian2025egor1} trains an RL agent for dynamic
tool calling over week-long recordings. HEM-LLM~\cite{cheng2024hemllm}
partitions videos at minima of adjacent-frame visual similarity
and compresses events into local and global visual memories.
Video-EM~\cite{wang2026videoem} groups query-relevant keyframes
into episodic events using visual temporal discontinuities and
temporal proximity.

Despite their diversity, memory construction remains a bottleneck.
Predefined temporal units can fragment coherent events or dilute
brief ones, while visual or visual--text signals can miss
speech-led transitions, confuse repeated instances, or omit
fleeting cues. Later retrieval must operate over these incomplete
units and cannot fully recover the lost event structure.

\section{Methodology}

\subsection{Overview Architecture}

We propose EMMA, a multimodal memory framework for long video
understanding built around explicit semantic event detection. The system
operates in three stages. (1)~\textbf{Event segmentation:} a dual-stream
detector locates semantic event boundaries by fusing visual and linguistic
signals through cross-modal validation. (2)~\textbf{Memory construction:}
detected boundaries organize video content into a three-level memory
hierarchy of raw video, Clip-level Memory (CLM), and Event-level Memory
(ELM). (3)~\textbf{Memory retrieval:} a multi-agent framework iteratively
queries this hierarchy to answer questions. As shown in
Figure~\ref{fig:overview}, the pipeline proceeds from left to right:
the dual-stream detector locates boundaries; detected boundaries
aggregate CLMs into semantically self-contained ELMs; and four
specialized agents collaboratively retrieve evidence from coarse events
to precise clips. We detail each stage below.

\subsection{Dual-Stream Event Segmentation}

A core challenge in building semantically coherent memory is deciding
where one event ends and another begins. Because a boundary may be marked
visually or verbally alone, a single-modality detector is structurally
insufficient. We therefore fuse a visual stream tracking semantic drift and
an ASR stream proposing chapter-level
candidates. Both streams are adjudicated through cross-modal validation
guided by an ASR reliability indicator $\rho$. This design maximizes boundary recall as each stream covers what the other misses, while a one-step look-ahead preserves precision.

\paragraph{Visual Stream.}

The visual stream maintains a Rolling Event Table (ET), a dynamically
updated record that accumulates the semantic narrative of the current
event as new clips arrive. Upon receiving a new CLM $c_t$, an LLM-based
updater $F_{\text{ET}}$ refreshes the ET and emits a transition signal:
\begin{equation}
\delta_t = F_{\text{ET}}(\text{ET}_{t-1}, c_t) \in \{\textsc{cont},\, \textsc{shift}\},
\end{equation}
where \textsc{cont} indicates $c_t$ continues the current event and
\textsc{shift} signals a potential boundary. The ET resets upon each
confirmed boundary, ensuring it reflects only the ongoing event. A
minimum cooldown interval between consecutive splits suppresses spurious
visual-only triggers immediately after a confirmed boundary.

\paragraph{ASR Stream.}

The ASR stream converts the video's audio into structured linguistic
signals. ASR transcribes spoken audio into
timestamped text transcripts. The resulting transcript
$\mathcal{T} = \{(s_i, e_i, w_i)\}_{i=1}^{N}$, where $s_i$ and $e_i$
denote the start and end timestamps and $w_i$ the text content of the
$i$-th utterance, is passed to an LLM
producing chapter-level boundary timestamps $\mathcal{B}_{\text{ASR}}$
and an LLM-produced reliability indicator
$\rho \in \{\textsc{High}, \textsc{Low}\}$ reflecting whether the
ASR transcript contains clear chapter-level structure.
The LLM assesses linguistic coherence and the
presence of explicit transition cues in the transcript; if the
transcript exhibits reliable chapter markers, the LLM sets
$\rho = \textsc{high}$, otherwise $\textsc{low}$. For each CLM $c_t$ spanning $[t_s, t_e)$, the ASR stream
emits a boundary indicator:
\begin{equation}
    \phi_t = \left[
        \exists\, b_k \in \mathcal{B}_{\text{ASR}}
        \;\text{s.t.}\; t_s \leq b_k < t_e
    \right],
\end{equation}
where $\phi_t = 1$ if a chapter boundary falls within the clip, and
$0$ otherwise. Both streams propose boundary candidates independently;
$\rho$ is forwarded to the validator below as a soft prior on how to
weigh their evidence, not as a hard switch.

\paragraph{Cross-Modal Boundary Confirmation.}

When either stream signals a boundary ($\delta_t = \textsc{shift}$ or
$\phi_t = 1$), the system enters a Pending state: it buffers the
current clip $c_t$ and inspects the next clip $c_{t+1}$ before
committing. This one-step look-ahead filters transient noise. If
$c_{t+1}$ returns to the current event theme, the signal is discarded
as noise and the event continues. If $c_{t+1}$ also diverges, a true
boundary is likely and the LLM-based validator $F_{\text{val}}$
proceeds to adjudication.

The validator jointly analyzes the current event state
$\text{ET}_{t-1}$, the two clips $c_t$ and $c_{t+1}$, and the combined
dual-stream evidence from $\delta_t$ and $\phi_t $ to produce a boundary decision
$v_t$ and, when applicable, a precise boundary timestamp $\tau^*$:
\begin{equation}
(v_t, \tau^*) = F_{\text{val}}(\text{ET}_{t-1}, c_t, c_{t+1}, \delta_{t}, \phi_{t}, \rho).
\end{equation}
The reliability indicator $\rho$ acts as a soft prior that biases the
validator toward the more trustworthy stream: when
$\rho{=}\textsc{high}$ the validator weighs the ASR signal more
heavily; when $\rho{=}\textsc{low}$ it relies more on visual evidence.
The validator outputs $v_t \in \{\textsc{keep}, \textsc{split}\}$.
If $v_t = \textsc{keep}$, the signal is rejected as transient noise
and the current event continues uninterrupted. If $v_t = \textsc{split}$,
the validator identifies the utterance in the ASR transcript that
marks the new event onset and uses its start timestamp as the
precise boundary $\tau^*$, falling back to $t_s(c_t)$ when ASR
is unavailable. CLMs preceding $\tau^*$ are consolidated into
an ELM (§3.4).


\subsection{Hierarchical Memory Construction}

Based on event boundaries produced by the dual-stream
segmentation module, we construct a three-level memory hierarchy
comprising raw video, Clip-level Memory (CLM), and Event-level
Memory (ELM), each serving a distinct role in balancing detail
preservation with semantic abstraction.

\paragraph{Raw Video.}

Raw video preserves the original uncompressed footage.
It is accessed on demand after a relevant time window
has been localized through CLM or ELM retrieval,
allowing a VLM to recover fine-grained visual details
that are lost during caption-based compression.

\paragraph{Clip-level Memory (CLM).}

CLM is the finest-grained indexed unit. Each CLM
corresponds to a $\tau_{\text{clip}}$-second video
segment with boundaries softly aligned to ASR
timestamps. A VLM processes frames at 1\,fps and
produces:
\begin{equation}
    \text{CLM}_t = (S_t, C_t, A_t, [t_s, t_e], \mathbf{e}_t),
\end{equation}
where $S_t$ is a summary, $C_t$ a visual caption,
$A_t$ the ASR transcript, $[t_s, t_e]$ the temporal
span, and $\mathbf{e}_t$ the embedding of all fields.
The Event Table (§3.2) is fed as context during
VLM generation to maintain narrative coherence
across clips.

\paragraph{Event-level Memory (ELM).}

ELM represents a complete semantic event. When a boundary
is confirmed, all CLMs since the previous boundary are
consolidated into one ELM via LLM summarization:
\begin{equation}
    \text{ELM}_k = (S_k, D_k, I_k, [t^k_s, t^k_e]),
\end{equation}
where $S_k$ is the event summary, $D_k$ a detailed
event description, and $I_k$ the reasoning
inferred from the event,  $[t^k_s, t^k_e]$ the span. 

Unlike CLM, which records fragmented
observations in fixed windows, ELM distills a more
concrete event morphology, enabling the memory retrieval layer to retrieve event-level gist
without scanning every CLM.

\subsection{Multi-Agent Collaborative Retrieval}

A long-video question often requires evidence from multiple
temporally distant events, and the right granularity (event
gist, clip caption, or raw footage) is unknown in advance.
A single retrieval round is insufficient because answers
frequently hinge on cross-event reasoning, and a monolithic
LLM cannot simultaneously plan, search, and critically
validate evidence. We therefore decompose retrieval into
an iterative loop shared by four specialized agents.

\paragraph{Query Agent.} Plans each round by selecting
the target memory level (ELM, CLM, or Raw) and formulating
a search query. It follows a broad-to-narrow strategy:
early rounds query ELM for coarse event localization;
later rounds query CLM or Raw for fine-grained detail.

\paragraph{Search Agent.} Executes retrieval against the
chosen memory level. For CLM and ELM, it scores each
candidate $m$ against query $q$ by combining semantic embedding similarity with keyword-level Jaccard overlap:
\begin{equation}
\text{score}(q,m) = \alpha \cos(\mathbf{e}_q,\mathbf{e}_m)
+ (1-\alpha)\, \text{Jaccard}(q,m),
\end{equation}
with weight $\alpha$. For Raw video, it invokes a VLM to
directly observe the localized segment and judge relevance.

\paragraph{Validate Agent.} Filters retrieved candidates
for relevance, merges them into an evidence pool, and
checks whether the accumulated evidence is sufficient
to answer the question. If not, it reports missing
information to the Query Agent to guide the next round.

\paragraph{Answer Agent.} Synthesizes the final answer
from the accumulated evidence pool once the validator
deems it sufficient, outputting the predicted option
with a supporting reasoning trace.

The four agents operate in a shared control loop.
The Query Agent plans; the Search Agent executes;
the Validate Agent assesses and decides whether to
continue; the Answer Agent terminates the loop by
producing the final answer. The loop exits when
evidence is sufficient or a maximum of rounds
is reached.

\section{Experiments}

\subsection{Experimental Setup}
\label{sec:setup}

\paragraph{Datasets and Metrics.}
We evaluate on three multiple-choice benchmarks that differ
in video genre, duration, and question type. Accuracy is
reported throughout following official protocols.
\begin{itemize}
\item \textbf{VideoMME-long}~\cite{fu2025videomme} is a general
long-video understanding benchmark with 300 hour-level videos
(avg.\ 41.4\,min) from diverse domains and 900 questions.

\item \textbf{LVBench}~\cite{wang2025lvbench} is an adversarial
long-content benchmark for structured reasoning, with 103 ultra-long
videos (avg.\ 4,101\,s, $\sim$68\,min) and 1,549 questions across six
categories: Entity Recognition (ER), Event Understanding (EU),
Key Information Retrieval (KIR), Reasoning (Rea), Temporal Grounding
(TG), and Summarization (Sum).

\item \textbf{EgoLifeQA}~\cite{yang2025egolife} is a first-person
persistent-memory benchmark built on week-long egocentric recordings
(avg.\ 44.3\,h) with 500 questions on daily activities across five
categories: EntityLog (EL), EventRecall (ER), HabitInsight (HI),
RelationMap (RM), and TaskMaster (TM).

\end{itemize}

\paragraph{Baseline Method.}
To assess EMMA against the full spectrum of existing
approaches, we compare four method categories that
progress from raw model capacity to structured memory
systems. Results for the first three categories are from
official leaderboards or original papers; memory-based
competitors are re-implemented under our unified backbone
(Qwen3-VL-8B + DeepSeek-R1~\cite{guo2025deepseek}) to ensure fair comparison.

\begin{itemize}

    \item \textbf{Proprietary models.}
We include GPT-4o~\cite{hurst2024gpt}, GPT-5~\cite{singh2025openai}, Gemini-1.5-Pro~\cite{team2024gemini}, and Gemini-2.5-Pro~\cite{comanici2025gemini} as
closed-source reference models. Their benchmark results are taken
from official leaderboards or prior reports, as summarized in
Table~\ref{tab:main}.

    \item \textbf{Open-source MLLMs.} Qwen2-VL~\cite{wang2024qwen2vl}, LLaVA-OV~\cite{li2024llavaonevision},
    InternVL2.5~\cite{chen2024internvl25}, and Qwen3-VL-8B~\cite{bai2025qwen3} uniformly sample frames
    within their context windows. AdaReTaKe~\cite{wang2025adaretake} reduces visual-token
    redundancy and VideoChat-Flash~\cite{li2024videochatflash} applies hierarchical visual
    compression. These methods scale by enlarging context, not
    by restructuring memory.
    \item \textbf{RAG-based methods.} Video-RAG~\cite{luo2024videorag} indexes auxiliary
    text for single-pass retrieval. LightRAG~\cite{guo2024lightrag} uses a dual-level
    graph combining entity and summary nodes. EgoRAG~\cite{yang2025egolife} is the
    EgoLife baseline. These methods test flat retrieval against
    our structured multi-round traversal.
    \item \textbf{Memory-based methods.} M3-Agent~\cite{long2025m3agent} (entity graph),
    VideoLucy~\cite{zuo2025videolucy} (hierarchical caption backtracking), and WorldMM~\cite{yeo2025worldmm}
    (multi-scale episodic memory). These are our direct
    competitors that maintain higher-level memory structures
    without explicit event detection at the base layer.
\end{itemize}

Our framework EMMA differs from all four categories. Unlike
full-context MLLMs, EMMA does not scale by enlarging the
context window but by restructuring video into an
event-aligned memory hierarchy. Unlike RAG systems, it
performs multi-round iterative retrieval across three
memory levels rather than single-pass flat retrieval.
Unlike existing memory-based agents, it explicitly detects
semantic event boundaries at the base layer, making memory
units aligned with natural event structure rather than
arbitrary temporal windows or entity graphs.

\paragraph{Implementation Details.}
We use Qwen3-VL-8B~\cite{bai2025qwen3} as the visual encoder for
clip captioning and visual observation, sampling video frames at
1\,fps;
ASR
transcripts are extracted from the audio track of each video.
DeepSeek-R1 serves as the LLM backbone for all reasoning
modules, including Event Table updates, boundary adjudication,
ELM summarization, and the four retrieval agents. We set the
base clip duration $\tau_{\text{clip}}=30$\,s, the minimum
inter-split interval $30$\,s (i.e., one clip),
the hybrid retrieval weight $\alpha=0.7$, the maximum retrieval
rounds $R=5$, and top-$k=5$ per round. No task-specific
fine-tuning is applied. To verify the
contribution of our event-driven memory system, all memory-based competitors (M3-Agent, WorldMM, VideoLucy) are re-implemented under the same Qwen3-VL-8B + DeepSeek-R1 backbone, while their
original memory construction and retrieval logic are preserved.
Code, prompt templates, and reproduction scripts are available at
\url{https://github.com/BAI-LAB/EMMA}.

\subsection{Main Results}

\begin{table}[t]
\centering
\caption{Overall accuracy (\%) on VideoMME-long, LVBench, and
EgoLifeQA. Memory-based systems use a unified Qwen3-VL-8B +
DeepSeek-R1 backbone; results for all other methods are taken from
official leaderboards or the corresponding papers.}
\label{tab:main}
\resizebox{\columnwidth}{!}{
\begin{tabular}{lccc}
\toprule
\textbf{Method} & \textbf{VideoMME-L} &
\textbf{LVBench} & \textbf{EgoLifeQA} \\
\midrule
\rowcolor{headerblue}
\multicolumn{4}{l}{\textit{Proprietary Models}} \\
GPT-4o                & 65.3 & 48.9 & 36.2 \\
GPT-5                 & 74.3 & 60.4 & 48.6 \\
Gemini-1.5-Pro        & 67.4 & 33.1 & 36.9 \\
Gemini-2.5-Pro        & 55.7 & 57.0 & 46.4 \\
\midrule
\rowcolor{headerblue}
\multicolumn{4}{l}{\textit{Open-source MLLMs}} \\
Qwen2-VL-72B        & 62.2 & 41.3 & --   \\
AdaReTaKe-72B       & 65.0 & 53.3 & --   \\
LLaVA-OV-72B        & 60.0 & --   & 30.8 \\
Qwen3-VL-8B         & 58.4 & 48.3 & 38.6 \\
InternVL2.5-78B     & 62.6 & 43.6 & --   \\
VideoChat-Flash-7B  & 55.6 & --   & 34.2 \\
\midrule
\rowcolor{headerblue}
\multicolumn{4}{l}{\textit{RAG-based Systems}} \\
LightRAG   & 46.6 & 30.4 & 48.8 \\
Video-RAG  & 60.1 & 33.1 & 55.4 \\
EgoRAG     & 41.1 & 32.2 & 36.0 \\
\midrule
\rowcolor{headerblue}
\multicolumn{4}{l}{\textit{Memory-based Systems}} \\
M3-Agent   & 57.8 & 51.5 & 55.6 \\
WorldMM    & 66.7 & 54.2 & 59.6 \\
VideoLucy  & 63.7 & 56.3 & 53.4 \\
\midrule
\rowcolor{ourspink}
\textbf{EMMA} & \textbf{71.0} &
\textbf{59.5} & \textbf{62.4} \\
\bottomrule
\end{tabular}}
\end{table}

\begin{table}[t]
\centering
\caption{Per-category accuracy (\%) on LVBench dataset.}

\label{tab:lvbench_detail}
\resizebox{\columnwidth}{!}{
\begin{tabular}{lccccccc}
\toprule
\textbf{Method} & \textbf{ER} & \textbf{EU} & 
\textbf{KIR} & \textbf{Rea} & \textbf{TG} & 
\textbf{Sum} & \textbf{Overall} \\
\midrule
GPT-4o          & 48.9 & 49.5 & 48.1 & 50.3 & 40.9 & 50.0 & 48.9 \\
Gemini 1.5 Pro  & 32.1 & 30.9 & 39.3 & 27.0 & 31.8 & 32.8 & 33.1 \\
Qwen2-VL-72B    & 38.0 & 41.1 & 38.3 & 46.5 & 41.4 & 46.6 & 41.3 \\
AdaReTaKe-72B   & 53.0 & 50.7 & 62.2 & 54.7 & 45.5 & 37.9 & 53.3 \\
InternVL2.5-78B & 43.8 & 42.0 & 42.1 & 51.0 & 36.8 & 37.9 & 43.6 \\
VideoAgent      & 28.0 & 30.3 & 28.0 & 28.0 & 29.3 & 36.4 & 29.3 \\
M3-Agent        & 47.8 & 52.9 & 60.2 & 51.0 & 47.7 & 50.9 & 51.5 \\
WorldMM         & 48.1 & 53.5 & 59.1 & 50.2 & 58.2 & 55.2 & 54.2 \\
VideoLucy       & 48.8 & 59.7 & 63.9 & 48.3 & 50.9 & 55.2 & 56.3 \\
\rowcolor{ourspink}
\textbf{EMMA}   & \textbf{56.4} & \textbf{60.9} & 
\textbf{64.9} & \textbf{54.2} & \textbf{60.5} & 
\textbf{56.9} & \textbf{59.5} \\
\bottomrule
\end{tabular}}
\end{table}

\begin{table}[t]
\centering
\caption{Per-category accuracy (\%) on EgoLifeQA dataset.}
\label{tab:egolifeqa_detail}
\resizebox{\columnwidth}{!}{
\begin{tabular}{lcccccc}
\toprule
\textbf{Method} & \textbf{EL} & \textbf{ER} & 
\textbf{HI} & \textbf{RM} & \textbf{TM} & 
\textbf{Avg.} \\
\midrule
GPT-4o         & 34.4 & 42.1 & 29.5 & 30.4 & 44.4 & 36.2 \\
Gemini 1.5 Pro & 36.0 & 37.3 & 45.9 & 30.4 & 34.9 & 36.9 \\
LLaVA-OV-72B   & 36.8 & 34.9 & 31.1 & 22.4 & 28.6 & 30.8 \\
EgoRAG         & 39.2 & 36.5 & 31.1 & 33.6 & 39.7 & 36.0 \\
M3-Agent       & 46.0 & 54.8 & 63.9 & 59.7 & 60.3 & 55.6 \\
VideoLucy      & 44.4 & 56.3 & 59.0 & 54.8 & 57.1 & 53.4 \\
WorldMM        & 51.6 & 58.7 & \textbf{68.9} & 58.9 & \textbf{69.8} & 59.6 \\
\rowcolor{ourspink}
\textbf{EMMA}  & \textbf{52.0} & \textbf{62.7} & 
67.2 & \textbf{66.4} & \textbf{69.8} & 
\textbf{62.4} \\
\bottomrule
\end{tabular}}
\end{table}

Table~\ref{tab:main} reports overall accuracy on VideoMME-long,
LVBench, and EgoLifeQA. For methods whose category-level results
are publicly available, Table~\ref{tab:lvbench_detail} and
Table~\ref{tab:egolifeqa_detail} provide per-category results on
LVBench and EgoLifeQA, covering question types from point-level
retrieval to event-level reasoning. We make four observations.

(1) EMMA consistently outperforms all compared memory-based systems
under the unified Qwen3-VL-8B + DeepSeek-R1 setting. Compared with
the strongest memory-based baseline on each benchmark, EMMA improves
by +4.3, +3.2, and +2.8 points on VideoMME-long, LVBench, and
EgoLifeQA, respectively. These gains indicate that event extraction,
hierarchical memory, and multi-agent retrieval together yield a
stronger design than existing memory organizations. EMMA is also
competitive with current proprietary models and achieves the highest
unweighted average performance among methods reporting results on
all three benchmarks.

(2) Hierarchical memory provides benefits beyond context-window
enlargement and caption-based retrieval. Compared with the strongest
open-source MLLM on each benchmark, EMMA improves by +6.0, +6.2,
and +23.8 points on VideoMME-long, LVBench, and EgoLifeQA,
respectively. It also surpasses the strongest RAG-based system by
+10.9, +26.4, and +7.0 points. These results suggest that queryable
multimodal memory becomes increasingly important as video duration
extends, while simply enlarging the context window or retrieving
caption-based segments remains insufficient.

(3) Event-driven memory units strengthen memory organization.
Existing memory-based systems maintain higher-level structures, such
as entity graphs, hierarchical captions, or multi-scale episodic
banks. EMMA additionally constructs its base memory units from
detected semantic events. The consistent gains in Table~\ref{tab:main},
together with the component and segmentation analyses in
Section~\ref{sec:ablation}, suggest that event-aligned cuts are an
important factor for memory organization effectiveness when backbone
capacity is fixed.

(4) Gains concentrate on event-structured questions. On questions
that require reasoning across multiple clips of the same coherent
event, as opposed to locating a single moment, EMMA shows
consistently positive gains over the strongest memory-based baseline,
with the largest reaching +6.7 on EgoLifeQA RM. On question types
that only require point-level retrieval (KIR and EL) or pattern
aggregation (HI), the effect is mixed. This selectivity is consistent
with the improvements arising from the event-alignment mechanism,
rather than backbone capacity or incidental data effects.

\begin{figure*}[t]
    \centering
    \begin{subfigure}[t]{0.49\textwidth}
        \centering
        \includegraphics[width=0.9\linewidth]{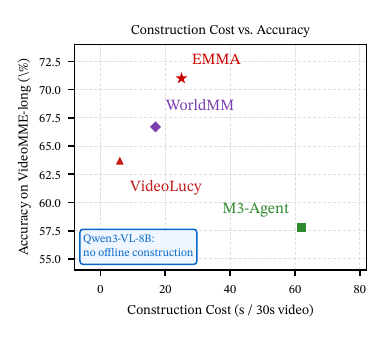}
        \caption{Accuracy v.s. Memory construction latency.}
        \label{fig:construction}
    \end{subfigure}
    \hfill
    \begin{subfigure}[t]{0.49\textwidth}
        \centering
        \includegraphics[width=0.9\linewidth]{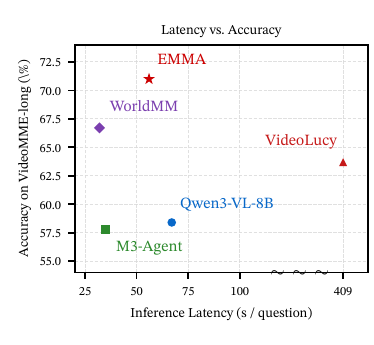}
        \caption{Accuracy v.s. Inference latency.}
        \label{fig:latency}
    \end{subfigure}
    \caption{Accuracy-efficiency trade-off on VideoMME-long. (a)~Offline memory construction latency per 30\,s of video. (b)~Per-question inference latency.}
    \label{fig:efficiency}
\end{figure*}

\subsection{Ablation Study}
\label{sec:ablation}

We study EMMA through component ablations and targeted analyses
of its event segmentation module.

\paragraph{Component contribution.}
To verify the contribution of each component in EMMA, we remove
one module at a time and measure the resulting drop in accuracy
and change in retrieval rounds. From
Table~\ref{tab:ablation}, we make three observations.
(1) Replacing dual-stream event segmentation with fixed-window
splitting produces the largest accuracy drop (-7.4, -2.2, and
-7.2 points) and requires more retrieval rounds, indicating that
event-aligned units provide more retrieval-friendly memory than
arbitrary windows. (2) The three memory levels are useful in
different ways: removing raw video mostly hurts when direct visual
evidence is needed, explaining its limited effect on LVBench;
removing CLM loses fine-grained evidence; and removing ELM lowers
accuracy while increasing retrieval rounds (2.03 vs.\ 1.68 on
VideoMME-long), showing that ELM provides efficient
coarse-level localization. (3) Replacing the multi-agent
retrieval loop with a single LLM call decreases performance by
-1.8, -2.3, and -9.0 points, respectively, showing the value of
iterative planning, search, and validation over a single-call
alternative.
A complementary single-agent multi-round control is reported in
Appendix~\ref{app:additional-eval}.
\begin{table}[t]
\centering
\caption{Component ablation on three benchmarks. Acc: accuracy
(\%); Turn: average retrieval rounds.}
\label{tab:ablation}
\resizebox{\columnwidth}{!}{
\begin{tabular}{lcccccc}
\toprule
\multirow{2}{*}{\textbf{Method}}
& \multicolumn{2}{c}{\textbf{VideoMME-L}}
& \multicolumn{2}{c}{\textbf{LVBench}}
& \multicolumn{2}{c}{\textbf{EgoLifeQA}} \\
\cmidrule(lr){2-3}\cmidrule(lr){4-5}\cmidrule(lr){6-7}
& Acc & Turn & Acc & Turn & Acc & Turn \\
\midrule
Fixed Window  & 63.6 & 2.26 & 57.3 & 2.95 & 55.2 & 3.60 \\
w/o Raw Video & 70.0 & 1.66 & 59.3 & 3.39 & 59.8 & 3.11 \\
w/o CLM       & 67.8 & 1.69 & 54.4 & 2.11 & 54.2 & 2.94 \\
w/o ELM       & 67.2 & 2.03 & 55.3 & 2.77 & 56.8 & 3.52 \\
w/o Multi-Agent & 69.2 & -- & 57.2 & -- & 53.4 & -- \\
\midrule
\rowcolor{ourspink}
\textbf{Full} & \textbf{71.0} & 1.68
              & \textbf{59.5} & 2.71
              & \textbf{62.4} & 3.16 \\
\bottomrule
\end{tabular}}
\end{table}

\paragraph{Event segmentation quality.}
The Fixed Window ablation above shows the downstream value of
event-aligned memory. We further directly evaluate whether EMMA
identifies boundaries that agree with external event references.
Using creator-provided chapter timestamps from 66 VideoMME-long
videos, we report boundary precision, recall, and F1 under 30\,s
and 60\,s tolerances, together with WindowDiff (WD) for overall
segmentation consistency. As shown in
Table~\ref{tab:segmentation_quality}, EMMA consistently
outperforms fixed-window baselines. Compared with the strongest
fixed-window result, EMMA improves F1@30 by 29.2 points and
F1@60 by 15.2 points, while achieving the lowest WD. This
improvement is also reflected in downstream QA on the same subset,
where EMMA reaches 73.23\% accuracy compared with 70.71\% for
Fixed 3min. Detailed evaluation protocols are provided in
Appendix~\ref{app:additional-eval}.

\begin{table*}[t]
\centering
\small
\caption{Event segmentation quality on 66 VideoMME-long videos
with creator-provided chapter timestamps as external reference
boundaries. P/R/F1 are reported under 30\,s and 60\,s
tolerances; lower WD is better.}
\label{tab:segmentation_quality}
\begin{tabular}{lccccccc}
\toprule
\textbf{Method}
& \textbf{P@30}$\uparrow$
& \textbf{R@30}$\uparrow$
& \textbf{F1@30}$\uparrow$
& \textbf{P@60}$\uparrow$
& \textbf{R@60}$\uparrow$
& \textbf{F1@60}$\uparrow$
& \textbf{WD}$\downarrow$ \\
\midrule
Fixed 3min  & 28.2 & 32.1 & 26.8 & 48.1 & 57.3 & 47.4 & 58.9 \\
Fixed 5min  & 30.5 & 19.2 & 21.1 & 45.9 & 30.7 & 33.5 & 51.5 \\
Fixed 10min & 31.2 & 9.2  & 13.0 & 45.0 & 14.5 & 20.3 & 45.6 \\
\midrule
\rowcolor{ourspink}
\textbf{EMMA}
& \textbf{57.3} & \textbf{67.2} & \textbf{56.0}
& \textbf{64.4} & \textbf{74.7} & \textbf{62.6}
& \textbf{38.9} \\
\bottomrule
\end{tabular}
\end{table*}

\paragraph{Dual-stream segmentation.}
To isolate the contribution of the two segmentation streams, we
evaluate a randomly sampled, fixed one-fifth subset of
VideoMME-long (180 questions), while keeping all other components
unchanged. Table~\ref{tab:dual_stream} shows that the dual-stream
setting achieves the highest accuracy on this controlled subset.
The visual stream captures semantic shifts that may not be
verbalized, whereas the ASR stream captures explicit topic
transitions.
The subset construction and controlled evaluation settings are
provided in Appendix~\ref{app:additional-eval}.
\begin{table}[t]
\centering
\small
\caption{Dual-stream segmentation ablation on a randomly sampled,
fixed 180-question subset of VideoMME-long. All non-segmentation
components follow the full EMMA configuration.}
\label{tab:dual_stream}
\begin{tabular}{lc}
\toprule
\textbf{Segmentation Stream} & \textbf{Accuracy (\%)} \\
\midrule
ASR only    & 71.11 \\
Visual only & 73.33 \\
\rowcolor{ourspink}
\textbf{Visual + ASR} & \textbf{76.11} \\
\bottomrule
\end{tabular}
\end{table}

\subsection{Efficiency Analysis}

Although EMMA introduces additional computation through event
segmentation, hierarchical memory construction, and multi-round
multi-agent retrieval, the cost remains moderate. We evaluate
this claim from two aspects. Figure~\ref{fig:construction}
measures offline memory construction cost per 30\,s of video:
EMMA requires $\sim$26\,s, lower than M3-Agent ($\sim$65\,s)
and moderately higher than WorldMM ($\sim$15\,s).
Figure~\ref{fig:latency} compares per-question inference
latency against accuracy: EMMA achieves the highest accuracy
(71\,\%) with a latency of $\sim$60\,s per question,
significantly performs Qwen3-VL-8B ($\sim$58\,\% at similar latency)
and substantially faster than VideoLucy, which pays a heavy
latency penalty for its iterative backtracking. 
Memory-based
methods such as WorldMM and M3-Agent show lower latency
but at the cost of significantly reduced accuracy.

Taken together, EMMA achieves the highest accuracy among the compared
memory-based systems while maintaining moderate construction and
inference costs. Relative to WorldMM, the fastest memory-based system
at inference, EMMA improves accuracy by 4.3 points at the cost of
additional latency. This trade-off suggests that the additional
computation introduced by hierarchical memory and multi-agent
retrieval is worthwhile for long-video understanding.

\subsection{Hyperparameter Analysis}

The choice of base clip duration $\tau_{\text{clip}}$ is
foundational to the memory hierarchy, as it determines the
granularity of the finest indexed layer. Finer clips provide
more detailed captions but increase retrieval complexity;
coarser clips reduce the search space but may lose
fine-grained evidence. To understand this trade-off, we
compare 10\,s and 30\,s clips on a 381-question LVBench
subset covering all six question types.

From Table~\ref{tab:granularity}, we can see that retrieval-bound categories (KIR, Sum) benefit from finer
clips, improving by +7.7 and +11.5 points respectively,
because shorter windows localize evidence more precisely.
Conversely, categories requiring event-level integration
(EU, Rea) degrade by -1.5 and -9.1 points under the same
setting, because overly fine segmentation disrupts the
coherence of event narratives. Entity Recognition remains
largely unchanged (+0.8). The opposing effects roughly
cancel on this subset (+1.6 overall).
These results validate a core design choice of our EMMA
architecture: no single granularity suits all question
types. Our three-level hierarchy resolves this tension
naturally. The 30\,s default is preferred because
holistic-reasoning categories dominate the full LVBench
distribution; when fine-grained detail is needed, CLM
provides clip-level captions within each event, and Raw
video recovers visual specifics on demand. The event-driven
ELM layer ensures that finer clips are always retrieved
within a coherent semantic unit, avoiding the fragmentation
that pure fixed-window splitting would cause.

\begin{table}[t]
\centering
\caption{Impact of clip duration on LVBench accuracy (\%).
A 381-question subset covers all six question types.}
\label{tab:granularity}
\resizebox{\columnwidth}{!}{
\begin{tabular}{lcccc}
\toprule
\textbf{Question Type} & \textbf{\#Q}
    & \textbf{30s CLM} & \textbf{10s CLM}
& \textbf{$\Delta$} \\
\midrule
Key Info.\ Retrieval  &  52 & 48.1 & 55.8 &  $+$7.7 \\
Summarization         &  26 & 50.0 & 61.5 & $+$11.5 \\
Temporal Grounding    &  15 & 60.0 & 80.0 & $+$20.0 \\
Entity Recognition    & 123 & 57.7 & 58.5 &  $+$0.8 \\
Event Understanding   & 132 & 61.4 & 59.8 &  $-$1.5 \\
Reasoning             &  33 & 57.6 & 48.5 &  $-$9.1 \\
\midrule
Overall               & 381 & 57.2 & 58.8 &  $+$1.6 \\
\bottomrule
\end{tabular}}
\end{table}

\section{Conclusion}
This paper argues that long-video understanding requires
memory units aligned with semantic events rather than
arbitrary temporal windows. We propose a multi-modal memory system EMMA, a framework
that detects event boundaries through cross-modal visual
and linguistic streams, organizes memory into a three-level
hierarchy of raw video, clips, and events, and retrieves
evidence through an iterative multi-agent loop. By grounding
memory construction in explicit event segmentation, EMMA
ensures that retrieval operates over semantically coherent
units rather than fragmented clips. This event-first
perspective offers a principled foundation for building
richer memory organizations in long-context multimodal
systems.

\section*{Limitations}

EMMA demonstrates that aligning memory units with semantic
events through dual-stream segmentation, a three-level
hierarchy, and multi-agent retrieval yields substantial
accuracy gains over memory-based baselines. Nevertheless,
the framework relies on multiple LLM calls during boundary
detection and ELM construction, which incurs higher
computational overhead than fixed-window approaches.
A natural direction is to distill the LLM-based validator
into a lightweight trained detector that runs in a single
forward pass. Additionally, the base clip granularity and
cross-modal boundary confirmation step length are currently
fixed globally; adapting these parameters to local content
density would likely close the residual gap further.
Although the one-step look-ahead validator suppresses most
false boundaries, segmentation errors can still propagate
to downstream retrieval, and a formal error-propagation
analysis is left to future work.

\section*{Acknowledgments}

This work was supported in part by the National Natural Science
Foundation of China (Grant No.~62236003).


\bibliography{references}

@inproceedings{fan2024videoagent,
  title={Videoagent: A memory-augmented multimodal agent for video understanding},
  author={Fan, Yue and Ma, Xiaojian and Wu, Rujie and Du, Yuntao and Li, Jiaqi and Gao, Zhi and Li, Qing},
  booktitle={European Conference on Computer Vision},
  pages={75--92},
  year={2024},
  organization={Springer}
}

@article{zuo2025videolucy,
  title={VideoLucy: Deep Memory Backtracking for Long Video Understanding},
  author={Zuo, Jialong and Deng, Yongtai and Kong, Lingdong and Yang, Jingkang and Jin, Rui and Zhang, Yiwei and Sang, Nong and Pan, Liang and Liu, Ziwei and Gao, Changxin},
  journal={Advances in Neural Information Processing Systems},
  volume={38},
  pages={25660--25691},
  year={2026}
}

@article{yeo2025worldmm,
  title={Worldmm: Dynamic multimodal memory agent for long video reasoning},
  author={Yeo, Woongyeong and Kim, Kangsan and Yoon, Jaehong and Hwang, Sung Ju},
  journal={arXiv preprint arXiv:2512.02425},
  year={2025}
}

@article{long2025m3agent,
  title={Seeing, listening, remembering, and reasoning: A multimodal agent with long-term memory},
  author={Long, Lin and He, Yichen and Ye, Wentao and Pan, Yiyuan and Lin, Yuan and Li, Hang and Zhao, Junbo and Li, Wei},
  journal={arXiv preprint arXiv:2508.09736},
  year={2025}
}

@article{tian2025egor1,
  title={Ego-r1: Chain-of-tool-thought for ultra-long egocentric video reasoning},
  author={Tian, Shulin and Wang, Ruiqi and Guo, Hongming and Wu, Penghao and Dong, Yuhao and Wang, Xiuying and Yang, Jingkang and Zhang, Hao and Zhu, Hongyuan and Liu, Ziwei},
  journal={arXiv preprint arXiv:2506.13654},
  year={2025}
}

@inproceedings{yang2025egolife,
  title={Egolife: Towards egocentric life assistant},
  author={Yang, Jingkang and Liu, Shuai and Guo, Hongming and Dong, Yuhao and Zhang, Xiamengwei and Zhang, Sicheng and Wang, Pengyun and Zhou, Zitang and Xie, Binzhu and Wang, Ziyue and others},
  booktitle={Proceedings of the Computer Vision and Pattern Recognition Conference},
  pages={28885--28900},
  year={2025}
}

@inproceedings{he2024malmm,
  title={Ma-lmm: Memory-augmented large multimodal model for long-term video understanding},
  author={He, Bo and Li, Hengduo and Jang, Young Kyun and Jia, Menglin and Cao, Xuefei and Shah, Ashish and Shrivastava, Abhinav and Lim, Ser-Nam},
  booktitle={Proceedings of the IEEE/CVF conference on computer vision and pattern recognition},
  pages={13504--13514},
  year={2024}
}

@article{zacks2007event,
  title={Event perception: a mind-brain perspective.},
  author={Zacks, Jeffrey M and Speer, Nicole K and Swallow, Khena M and Braver, Todd S and Reynolds, Jeremy R},
  journal={Psychological bulletin},
  volume={133},
  number={2},
  pages={273},
  year={2007},
  publisher={American Psychological Association}
}

@article{li2024videochatflash,
  title={Videochat-flash: Hierarchical compression for long-context video modeling},
  author={Li, Xinhao and Wang, Yi and Yu, Jiashuo and Zeng, Xiangyu and Zhu, Yuhan and Huang, Haian and Gao, Jianfei and Li, Kunchang and He, Yinan and Wang, Chenting and others},
  journal={arXiv preprint arXiv:2501.00574},
  year={2024}
}

@inproceedings{fu2025videomme,
  title={Video-MME: The First-Ever Comprehensive Evaluation Benchmark of Multi-modal {LLM}s in Video Analysis},
  author={Fu, Chaoyou and Dai, Yuhan and Luo, Yongdong and Li, Lei and Ren, Shuhuai and Zhang, Renrui and Wang, Zihan and Zhou, Chenyu and Shen, Yunhang and Zhang, Mengdan and others},
  booktitle={Proceedings of the IEEE/CVF Conference on Computer Vision and Pattern Recognition (CVPR)},
  pages={24108--24118},
  year={2025}
}

@inproceedings{wang2025lvbench,
  title={Lvbench: An extreme long video understanding benchmark},
  author={Wang, Weihan and He, Zehai and Hong, Wenyi and Cheng, Yean and Zhang, Xiaohan and Qi, Ji and Ding, Ming and Gu, Xiaotao and Huang, Shiyu and Xu, Bin and others},
  booktitle={Proceedings of the IEEE/CVF International Conference on Computer Vision},
  pages={22958--22967},
  year={2025}
}

@article{bai2025qwen3,
  title={Qwen3-{VL} technical report},
  author={Bai, Shuai and Cai, Yuxuan and Chen, Ruizhe and Chen, Keqin and Chen, Xionghui and Cheng, Zesen and Deng, Lianghao and Ding, Wei and Gao, Chang and Ge, Chunjiang and others},
  journal={arXiv preprint arXiv:2511.21631},
  year={2025}
}

@article{guo2025deepseek,
  title={DeepSeek-{R1}: Incentivizing Reasoning Capability in {LLM}s via Reinforcement Learning},
  author={Guo, Daya and Yang, Dejian and Zhang, Haowei and Song, Junxiao and Zhang, Ruoyu and Xu, Runxin and Zhu, Qihao and Ma, Shirong and Wang, Peiyi and Bi, Xiao and others},
  journal={arXiv preprint arXiv:2501.12948},
  year={2025}
}

@article{luo2024videorag,
  title={Video-rag: Visually-aligned retrieval-augmented long video comprehension},
  author={Luo, Yongdong and Zheng, Xiawu and Li, Guilin and Yin, Shukang and Lin, Haojia and Fu, Chaoyou and Huang, Jinfa and Ji, Jiayi and Chao, Fei and Luo, Jiebo and others},
  journal={Advances in Neural Information Processing Systems},
  volume={38},
  pages={168008--168033},
  year={2026}
}

@inproceedings{guo2024lightrag,
  title={LightRAG: Simple and Fast Retrieval-Augmented Generation},
  author={Guo, Zirui and Xia, Lianghao and Yu, Yanhua and Ao, Tu and Huang, Chao},
  booktitle={Findings of the Association for Computational Linguistics: EMNLP 2025},
  year={2025},
  url={https://aclanthology.org/2025.findings-emnlp.568/}
}

@article{wang2024qwen2vl,
  title={Qwen2-{VL}: Enhancing Vision-Language Model's Perception of the World at Any Resolution},
  author={Wang, Peng and Bai, Shuai and Tan, Sinan and Wang, Shijie and Fan, Zhihao and Bai, Jinze and Chen, Keqin and Liu, Xuejing and Wang, Jialin and Ge, Wenbin and others},
  journal={arXiv preprint arXiv:2409.12191},
  year={2024}
}

@inproceedings{wang2025adaretake,
  title={AdaReTaKe: Adaptive redundancy reduction to perceive longer for video-language understanding},
  author={Wang, Xiao and Si, Qingyi and Zhu, Shiyu and Wu, Jianlong and Cao, Li and Nie, Liqiang},
  booktitle={Findings of the Association for Computational Linguistics: ACL 2025},
  pages={5417--5432},
  year={2025}
}

@article{li2024llavaonevision,
  title={{LLaVA}-{O}ne{V}ision: Easy Visual Task Transfer},
  author={Li, Bo and Zhang, Yuanhan and Guo, Dong and Zhang, Renrui and Li, Feng and Zhang, Hao and Zhang, Kaichen and Zhang, Peiyuan and Li, Yanwei and Liu, Ziwei and others},
  journal={arXiv preprint arXiv:2408.03326},
  year={2024}
}

@article{chen2024internvl25,
  title={Expanding Performance Boundaries of Open-Source Multimodal Models with Model, Data, and Test-Time Scaling},
  author={Chen, Zhe and Wang, Weiyun and Cao, Yue and Liu, Yangzhou and Gao, Zhangwei and Cui, Erfei and Zhu, Jinguo and others},
  journal={arXiv preprint arXiv:2412.05271},
  year={2024}
}

@inproceedings{song2024moviechat,
  title={MovieChat: From Dense Token to Sparse Memory for Long Video Understanding},
  author={Song, Enxin and Chai, Wenhao and Wang, Guanhong and Zhang, Yucheng and Zhou, Haoyang and Wu, Feiyang and others},
  booktitle={Proceedings of the IEEE/CVF Conference on Computer Vision and Pattern Recognition (CVPR)},
  pages={18221--18232},
  year={2024}
}

@inproceedings{wang2024videotree,
  title={VideoTree: Adaptive Tree-Based Video Representation for {LLM} Reasoning on Long Videos},
  author={Wang, Ziyang and Yu, Shoubin and Stengel-Eskin, Elias and Yoon, Jaehong and Cheng, Feng and Bertasius, Gedas and Bansal, Mohit},
  booktitle={Proceedings of the IEEE/CVF Conference on Computer Vision and Pattern Recognition (CVPR)},
  year={2025}
}

@inproceedings{liu2023llava,
  title={Visual Instruction Tuning},
  author={Liu, Haotian and Li, Chunyuan and Wu, Qingyang and Lee, Yong Jae},
  booktitle={Advances in Neural Information Processing Systems (NeurIPS)},
  year={2023}
}

@article{team2024gemini,
  title={Gemini 1.5: Unlocking multimodal understanding across millions of tokens of context},
  author={Team, Gemini and Georgiev, Petko and Lei, Ving Ian and Burnell, Ryan and Bai, Libin and Gulati, Anmol and Tanzer, Garrett and Vincent, Damien and Pan, Zhufeng and Wang, Shibo and others},
  journal={arXiv preprint arXiv:2403.05530},
  year={2024}
}

@article{hurst2024gpt,
  title={Gpt-4o system card},
  author={Hurst, Aaron and Lerer, Adam and Goucher, Adam P and Perelman, Adam and Ramesh, Aditya and Clark, Aidan and Ostrow, AJ and Welihinda, Akila and Hayes, Alan and Radford, Alec and others},
  journal={arXiv preprint arXiv:2410.21276},
  year={2024}
}

@inproceedings{cheng2024hemllm,
  title={Enhancing long video understanding via hierarchical event-based memory},
  author={Cheng, Dingxin and Li, Mingda and Liu, Jingyu and Guo, Yongxin and Jiang, Bin and Liu, Qingbin and Chen, Xi and Zhao, Bo},
  booktitle={2025 IEEE International Conference on Multimedia and Expo (ICME)},
  pages={1--6},
  year={2025},
  organization={IEEE}
}

@inproceedings{zeng2025scenerag,
  title={Scenerag: Scene-level retrieval-augmented generation for video understanding},
  author={Zeng, Nianbo and Hou, Haowen and Yu, F Richard and Shi, Si and He, Ying Tiffany},
  booktitle={ICASSP 2026-2026 IEEE International Conference on Acoustics, Speech and Signal Processing (ICASSP)},
  pages={12122--12126},
  year={2026},
  organization={IEEE}
}

@article{wang2026videoem,
  title={Video-EM: Event-Centric Episodic Memory for Long-Form Video Understanding},
  author={Wang, Yun and Zhang, Long and Liu, Jingren and Yan, Jiaqi and Zhang, Zhanjie and Zheng, Jiahao and Ma, Ao and Ling, Run and Yang, Xun and Wu, Dapeng and others},
  journal={arXiv preprint arXiv:2508.09486},
  year={2025}
}

@article{singh2025openai,
  title={Openai gpt-5 system card},
  author={Singh, Aaditya and Fry, Adam and Perelman, Adam and Tart, Adam and Ganesh, Adi and El-Kishky, Ahmed and McLaughlin, Aidan and Low, Aiden and Ostrow, AJ and Ananthram, Akhila and others},
  journal={arXiv preprint arXiv:2601.03267},
  year={2025}
}

@article{comanici2025gemini,
  title={Gemini 2.5: Pushing the frontier with advanced reasoning, multimodality, long context, and next generation agentic capabilities},
  author={Comanici, Gheorghe and Bieber, Eric and Schaekermann, Mike and Pasupat, Ice and Sachdeva, Noveen and Dhillon, Inderjit and Blistein, Marcel and Ram, Ori and Zhang, Dan and Rosen, Evan and others},
  journal={arXiv preprint arXiv:2507.06261},
  year={2025}
}

\clearpage
\appendix

\onecolumn

\begin{center}
\includegraphics[width=\textwidth]{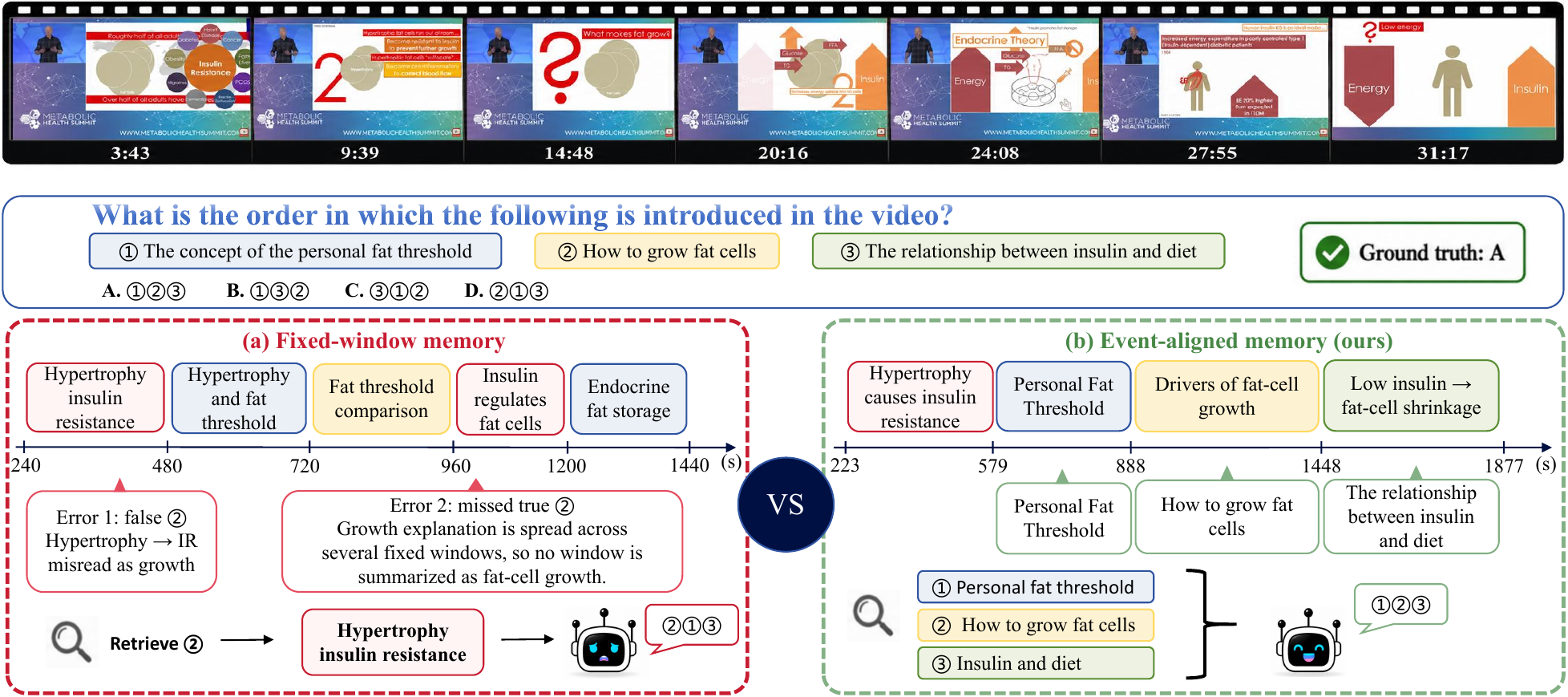}
\captionof{figure}{Lecture-video case requiring temporal ordering of three
    topics. \textbf{Left:} fixed-window memory mis-attributes
    topic~\protect\circled{1}'s hypertrophy mention to topic~\protect\circled{2}
    and fragments topic~\protect\circled{2}'s true content across three windows,
    yielding the wrong order. \textbf{Right:} event-aligned memory keeps
    each topic as one self-contained unit and recovers the correct order.}
\label{fig:case}
\end{center}

\vspace{1em}

\nolinenumbers
\appendix
\begin{multicols}{2}

\section{Case Study}
\label{app:case}

Figure~\ref{fig:case} grounds the fragmentation failure described in
Section~\ref{sec:intro} on a lecture-video question that asks for the
order of three topics. Two compounding errors trip up the fixed-window
system. First, a discussion of hypertrophy that belongs to
topic~\circled{1} (``what we eat $\to$ fat'') gets misread as evidence
for topic~\circled{2} (``how to grow fat cells''), because hypertrophy
is superficially associated with cell growth. Second, the actual
content of topic~\circled{2} is split across three subsequent
fixed-windows, so its true onset is hard to localize and the misleading
hypertrophy mention dominates retrieval. Event-aligned memory keeps
topic~\circled{1} and topic~\circled{2} as separate self-contained
units, so hypertrophy stays attached to its correct event and
topic~\circled{2}'s onset is recovered in a single retrieval, yielding
the correct order.

\section{Re-implementation of Memory-Based Baselines}
\label{app:reimpl}

To isolate the effect of memory organization, all three memory-based
competitors are re-evaluated under the same Qwen3-VL-8B + DeepSeek-R1
backbone used by our framework. We keep each method's memory
construction and retrieval logic intact and adapt only the model
interfaces.

\paragraph{M3-Agent and VideoLucy.}
Both systems are originally written against generic VLM/LLM
interfaces. We replace their visual encoders with Qwen3-VL-8B for
clip captioning and visual observation, and route all reasoning calls
(memory updates, retrieval planning, and answer generation) through
DeepSeek-R1. No other modifications are made to their public
implementations.

\paragraph{WorldMM.}
WorldMM additionally assumes a QA model that natively ingests
retrieved frames alongside text, which DeepSeek-R1 does not support.
We therefore adopt the same text-mediated visual interface used by
VideoLucy and by our own retrieval loop: when visual input is
required, the retrieved visual segment is first passed through
Qwen3-VL-8B together with the current question to produce a
query-specific description, which is then appended to DeepSeek-R1's
context. This adaptation preserves WorldMM's adaptive multi-scale
memory selection logic and changes only the visual-to-text bridging
layer.

\section{Evaluation Protocols and Additional Controls}
\label{app:additional-eval}

\subsection{Event Segmentation Evaluation Protocol}

We evaluate event segmentation on the 66 VideoMME-long videos that
provide creator-authored chapter timestamps. We treat these timestamps
as external reference boundaries; they are used only for evaluation
and are not available during memory construction or question
answering. Fixed-window baselines partition each full video at
uniform 3, 5, or 10 minute intervals.

We report boundary precision, recall, and F1 under tolerance windows
of 30\,s and 60\,s. These metrics quantify the agreement between
predicted and reference boundaries under the corresponding temporal
tolerance. We additionally report WindowDiff (WD), which measures
disagreement in whether temporally separated points belong to the
same segment; lower values indicate better agreement with the
reference segmentation. The detailed results are reported in
Table~\ref{tab:segmentation_quality}. For the downstream QA
comparison, we restrict evaluation to questions associated with the
same 66 videos.

\subsection{Controlled Retrieval Studies}

For both the dual-stream ablation in
Table~\ref{tab:dual_stream} and the following multi-agent control, we
use a randomly sampled and fixed one-fifth subset of VideoMME-long
containing 180 questions. All variants use identical questions and
keep the remaining memory construction and retrieval settings fixed.
The selected question IDs and sampling script are included in the
released code repository.

To separate the benefit of role specialization from that of iterative
retrieval, we replace EMMA's four specialized agents with a single
agent controller while retaining multi-round retrieval. As shown in
Table~\ref{tab:single_agent}, the specialized multi-agent design
improves accuracy by 7.22 points while requiring fewer retrieval
rounds.

\begin{center}
\small
\captionof{table}{Single-agent multi-round control on the fixed
180-question VideoMME-long subset.}
\label{tab:single_agent}
\resizebox{\columnwidth}{!}{
\begin{tabular}{lcc}
\toprule
\textbf{Method} & \textbf{Accuracy (\%)} &
\textbf{Avg. Rounds} \\
\midrule
Single-agent multi-round & 68.89 & 3.03 \\
\rowcolor{ourspink}
\textbf{EMMA (multi-agent multi-round)}
& \textbf{76.11} & \textbf{1.69} \\
\bottomrule
\end{tabular}}
\end{center}

\end{multicols}
\end{document}